 \documentclass[preprint]{elsarticle}
 \usepackage{aas_macros}

\usepackage{amssymb}
\usepackage{lipsum}
\usepackage{courier} 
\usepackage{multirow}

\journal{Astronomy $\&$ Computing}

\begin{document}

\begin{frontmatter}



\title{Morphological classification of eclipsing binary stars using computer vision methods}


\author[upjs]{\v{S}. Parimucha}
\author[upjs]{M. Gabdeev\corref{mycorrespondingauthor}}
\cortext[mycorrespondingauthor]{Corresponding author}

\author[upjs]{Y. Markus}
\author[ausav]{M. Va\v{n}ko}
\author[upjs,ond]{P. Gajdo\v{s}}

\address[upjs]{Institute of Physics, Faculty of Science, Pavol Jozef \v{S}af\'arik University in Ko\v{s}ice, Park Angelinum 9, 040 01 Ko\v{s}ice, Slovakia}

\address[ausav]{Astronomical Institute, Slovak Academy of Sciences, 059 60 Tatranská
Lomnica, Slovakia}

\address[ond]{Astronomical Institute, Czech Academy of Sciences, Fri\v{c}ova 298, 25 165 Ond\v{r}ejov, Czech Republic}

\begin{abstract}
We present an application of computer vision methods to classify the light curves of eclipsing binaries (EB). We have used pre-trained models based on convolutional neural networks (\textit{ResNet50}) and vision transformers (\textit{vit\_base\_patch16\_224}), which were fine-tuned on images created from synthetic datasets. To improve model generalisation and reduce overfitting, we developed a novel image representation by transforming phase-folded light curves into polar coordinates combined with hexbin visualisation. Our hierarchical approach in the first stage classifies systems into detached and overcontact types, and in the second stage identifies the presence or absence of spots. The binary classification models achieved high accuracy ($>96\%$) on validation data across multiple passbands (Gaia~$G$, $I$, and $TESS$) and demonstrated strong performance ($>94\%$, up to $100\%$ for $TESS$) when tested on extensive observational data from the OGLE, DEBCat, and WUMaCat catalogues. While the primary binary classification was highly successful, the secondary task of automated spot detection performed poorly, revealing a significant limitation of our models for identifying subtle photometric features. This study highlights the potential of computer vision for EB morphological classification in large-scale surveys, but underscores the need for further research into robust, automated spot detection.
\end{abstract}


\end{frontmatter}

\section{Introduction}
\label{introduction}

In recent years, several ground-based and space surveys have produced large volumes of photometric observations covering almost the entire sky. These observations improved the precision and temporal coverage of astronomical research and significantly transformed workflows of astronomical data analysis. Consequently, they have introduced new challenges in analysing  large numbers of different types light curves of variable stars. 

An important task in the study of the photometric variations of stars is their classification into distinct variability classes. \citet{2011ApJ...733...10R} introduced a method for variable star classification using a random forest classifier based on features extracted from light curves. Their approach effectively identified specific variability classes, such as pulsating and eclipsing variables.

\citet{2016A&A...587A..18K} introduced a machine learning package for the automated classification of periodic variable stars, trained on 16 features extracted from light curves. A deep learning model based on convolutional neural networks (CNNs) was published by \citet{2019MNRAS.482.5078A}.  Their architecture processed differences in time and magnitude of light curves, effectively capturing classification patterns irrespective of cadence and filter variations. \citet{2023A&A...674A..14R} classified variable sources from Gaia Data Release 3 using eXtreme Gradient Boosting and the Random Forest method into 25 classes, including EBs. Wang et al.\cite{2023arXiv230513745W} showed the effectiveness of computer vision techniques in classifying light curves from various variability types. Their study involved converting light curves into images using several algorithms including wavelet transformation. They demonstrated the potential of applying CNNs for accurate and efficient classification. Shan et al.\citep{2025PASP..137d4503S} recently developed a hybrid deep learning model, CALNet, which fuses light curve vectors with periodogram data to identify EBs in TESS observations with a high recall rate of 99.1\%.

All previously mentioned papers focused on the overall classification of variable stars. However, few have addressed to the specific detected subgroup of EBs. EBs can also be divided depending on the shapes of their light curves and/or their physical properties (see Section~\ref{sec:eclipsingbinaries}). Correct classification is crucial for a detailed analysis of these systems and for determining their orbital and stellar parameters. This task is not straightforward since it represents an inverse problem. All traditional packages for light curve analysis, e.g Phoebe \citep{prsa02}, ELC \citep{Orosz00}, or ELISa \citep{2021A&A...652A.156C}, require a lot of human interaction. The process is sensitive to the estimation of the initial values of parameters and depends on the morphological type of binary. 

Several authors have attempted to classify EBs into different groups. \citet{2006A&A...446..395S}  presented an automatic classification of EBs light curves into 7 types based on historical phenomenological characteristics (see Section~\ref{sec:eclipsingbinaries}). They used a Bayesian set of neural networks trained on data collected by the Hipparcos satellite.  On the other side, \citet{2021A&C....3600488C} used a combination of bidirectional Long Short-Term Memory (LSTM) and a one-dimensional CNN trained on synthetic light curves divided into two groups -- detached and contact. Each system was represented by a 100-point vector with a resulting accuracy of 98\% on the evaluation dataset. Following their work, \citet{2024CoSka..54b.167P} expanded the classification scheme to systems with and without spots and obtained 97\% accuracy. 

This paper focuses on classifying EBs into different morphological groups using a computer vision approach. In Section~\ref{sec:eclipsingbinaries}, we give short information about EBs and our classification scheme. In Section~\ref{sec:comp_vis}, the models used for computer vision are noted and we explain the transformation of light curves into images. In Section~\ref{sec:datasets}, we describe the datasets used for training, validation, and testing. In the last sections, we discuss the performance of our models. 

\section{Eclipsing binary stars}
\label{sec:eclipsingbinaries}

EBs belong to the group of variable stars, where changes in the system's geometry cause brightness variations. They are binary systems where the components move around a common centre of mass and mutually eclipse one another from the observer's line of sight \citep[e.g.,][]{percy2007understanding}. This produces typical light curves, where almost all information about components and their orbit is encoded \citep{prsa2019}.   

From the analysis of the light curves, we can determine the photometric parameters of EB, orbital inclination $i$, orbital eccentricity $e$, photometric mass ratio of the components $q=m_2/m_1$, the potentials of the components $\Omega_1$ and $\Omega_2$, which describe their shapes and effective temperature ratio of components $T_1/T_2$. By combining photometric parameters with spectroscopic parameters, determined from radial velocities, we can also derive absolute parameters, such as the components' masses, radii, and luminosities, as well as the mutual distance between stars in a system \citep{2001icbs.book.....H}. These parameters can be determined with a precision of up to a few percent \citep{2000A&A...362..169P}. Knowing the distance to a system from independent measurements (e.g., from Gaia) allows us to determine the absolute parameters without spectroscopic observations \citep{2023RMxAA..59..137K}. The precision of this method is worse than in the previous case (10--20\%). Nevertheless, it should be noted that for most EBs, light curve is the only source of information, and we will not be able to measure radial velocities because these systems are too faint to observe with high-resolution spectrographs, even on large telescopes. 

EBs were historically divided into three phenomenological groups based only on the shape of the observed light curves. Three classes can be distinguished:
\begin{itemize}
    \item {\bf Algol} ($\beta$~Per) -- light curves show plateaus in parts with maximum brightness and two significant brightness minima,
    \item {\bf $\beta$~Lyr} -- light curves show continual changes in measured brightness, and the minima have a different depth,
    \item {\bf W UMa} -- light curves show a continual change, similar to $\beta$~Lyr type, but the minima depths are similar.
\end{itemize}

 The given classification reflects the shapes of light curves and does not consider underlying physical characteristics hidden beyond the object of origin. 

The second possible classification was introduced by \citet{1959cbs..book.....K}. It is based on the concept of the Roche model, introduced by French mathematician \'{E}douard Roche in the 19$^{\rm th}$ century. It defines a region of space around each star within a binary system, where gravitational forces dominate, and the material is bound to that particular star \citep{2001icbs.book.....H}. \citet{1979ApJ...234.1054W} introduced the generalised Roche model and considered the components' elliptical orbit and non-synchronous rotation.  For the circular orbit and the synchronous components, we have three possible configurations:
\begin{itemize}
    \item {\bf detached (D)} -- both components  are inside their Roche lobes. If the components are small compared with their Roche lobes, their shapes closely approximate spheres,
    \item {\bf semi-detached (SD)} -- one of the components fills its lobe and mass can be transferred to another component through inner Lagrangian point L$_1$,
    \item {\bf contact} or {\bf overcontact (C)} -- each component has a surface larger than its Roche lobe, and a common envelope around both components with one common potential is created. 
\end{itemize}

The connection between the mentioned classifications may seem obvious at first glance; Algol types are detached systems, $\beta$~Lyr are semi-detached, and W~UMa stars are contact systems. But in reality, it is much more complicated. Many detached and contact systems have light curves that resemble $\beta$~Lyr type. On the other side, semi-detached systems can have Algol-type light curves. The Algol itself is an example of this. 

Semi-detached systems are, in fact, detached, because for the description of their shapes, we need in the Roche geometry two potentials $\Omega_1$ and $\Omega_2$, while for overcontact systems, the shapes of components are determined by only one common potential, $\Omega_1 = \Omega_2 = \Omega$. Such classification forms the foundational step in our larger, hierarchical research program. The ultimate goal of this program is to develop a deep learning framework capable of determining the physical parameters of binary systems directly from their light curves. To support this objective, the initial classification must be physically motivated to guide the subsequent parameter-fitting stage. 

For physical modelling, the most critical distinction is whether the two stars share one common potential (overcontact systems) or are described by two separate potentials. Since both detached and semi-detached systems fall into this latter category, we treat them as a single group. Therefore, our methodology focuses on making this primary, physically meaningful distinction between overcontact and non-contact systems.

Moreover, many EBs are affected by the presence of spots on the star surface. It is a common phenomenon in late-type stars. The spots cause changes in the light curves' brightness and their asymmetry. The most common observational consequence is the O'Connel effect, which refers to unequal brightness on maxima. The presence of spots affects the determination of photometric parameters and must be eliminated during the light curve analysis. This process is not straightforward because the position of the spot and its diameter cannot be derived from photometry alone \citep{prsa2019}.

Based on these facts, we decided to build our EBs classification into two main groups: detached and overcontact. We divided further each of these groups into two subgroups: systems with and without spots. Applying the correct categorization to data from large photometric surveys into these four groups can allow us to select appropriate procedures to analyze EBs, and determine absolute parameters for those systems. 

\section{Computer vision and image classification }
\label{sec:comp_vis}

Computer vision (CV) is a field of artificial intelligence (AI) that enables machines to interpret and understand visual data from the world. It can recognize image patterns, enabling image classification, image recognition, image segmentation, object detection, and video analysis techniques. We used image classification, where an algorithm assigns a label to an image based on its content. It typically involves deep learning models, such as CNNs \citep{he2016deep} and vision transformers (ViTs) \citep{dosovitskiy2021imageworth16x16words}, which learn to recognize patterns and features from large datasets.

Deep learning CV models, particularly CNNs and ViTs, offer robustness to noise and missing data, which is useful in analysing observational light curves of eclipsing binaries. CNNs extract spatial features from images hierarchically, allowing them to recognize patterns even with gaps or distortions. Data augmentation techniques during training, such as adding artificial noise and data removal, further enhance the model's generalisation. ViT leverages global self-attention mechanisms, making them less sensitive to locally missing values \citep{DBLP:journals/corr/abs-2105-10497}. Transfer learning from pre-trained models also improves classification accuracy, even with limited labelled datasets. These properties make image-based classification a powerful tool for analysing large astronomical datasets.

In this study, we chose to fine-tuned the following pre-trained models:
\begin{itemize}
    \item {\bf ResNet50} - CNN based architecture model \citep{he2016deep} known for its depth and efficiency in image classification tasks. It employs residual blocks, incorporating "skip connections" and allowing information to bypass some layers. This innovative design mitigates the vanishing/exploding gradient problem, enabling the training of deeper networks more effectively.
    
    \item {\bf vit\_base\_patch16\_224} - ViT model developed by Google, designed especially for image classification tasks \citep{dosovitskiy2021imageworth16x16words}. It is inspired by transformers in natural language processing. It treats images as sequences of patches and utilizes self-attention mechanisms to capture global dependencies. This approach has shown competitive performance on image recognition benchmarks, often surpassing CNNs in accuracy and efficiency when trained on sufficiently large datasets.
\end{itemize}

Both models are available with pre-trained weights, allowing efficient fine-tuning on our specific dataset, which is described in Section \ref{sec:models}. 

\subsection{Light curves transformation}
\label{sec:transformation}
A fundamental decision in applying machine learning to light curves is the choice of data representation: treating them as time-series vectors or as images. While vector representations are common, they present a significant challenge when dealing with sparse and non-uniformly sampled data, a key characteristic of surveys like Gaia.

Given that the Gaia archive contains the largest number of eclipsing binary candidates, building a tool applicable to its data was one of our primary objectives. Therefore, we adopted an image-based approach for this work. Our next step was to find the optimal method for transforming our light curve data into these images.

The obvious way is to create images with classical phase diagrams, which was also our first step. When we used them to train the models, we encountered problems with the classification model over-fitting to the training data. This suggests that the model was learning features specific to the training set rather than generalisable patterns inherent to the different classes of light curves.

To address this issue, we began an iterative search for a more robust image representation. We explored several alternative hypotheses, including: coding individual data points by colour and size based on their normalized flux values; a multimodal technique that attempted to incorporate the orbital period as an additional input feature. However, none of these approaches yielded the desired improvement in performance and generalisation.

\begin{figure}[t]
    \centering 
    \includegraphics[width=0.4\textwidth]{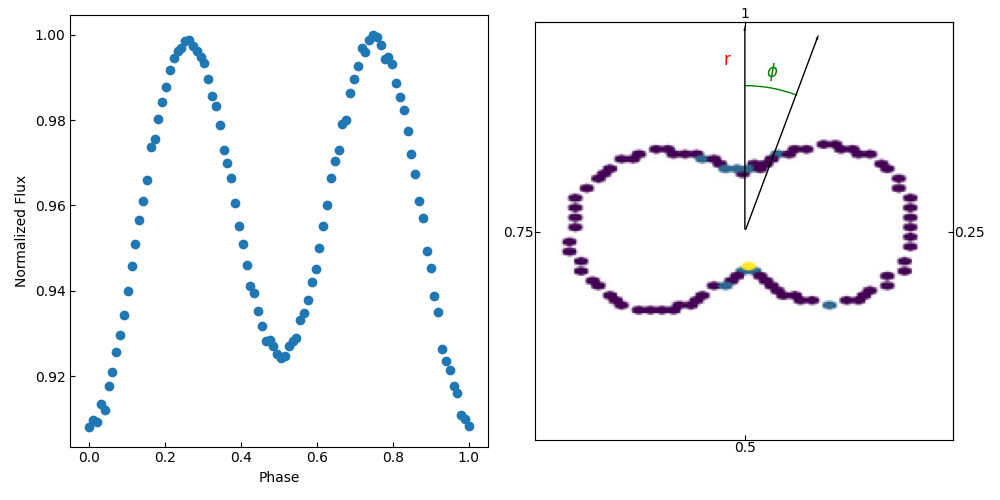}	
    \includegraphics[width=0.4\textwidth]{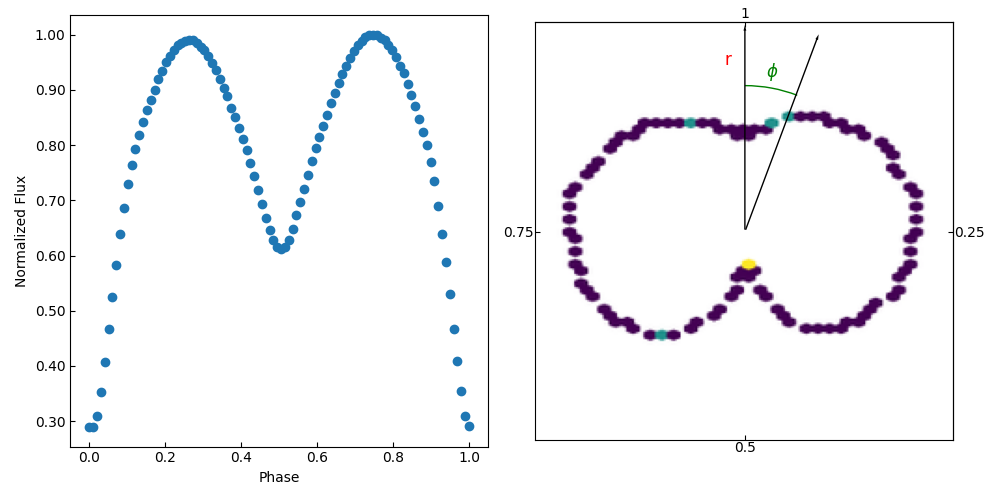}
    \includegraphics[width=0.4\textwidth]{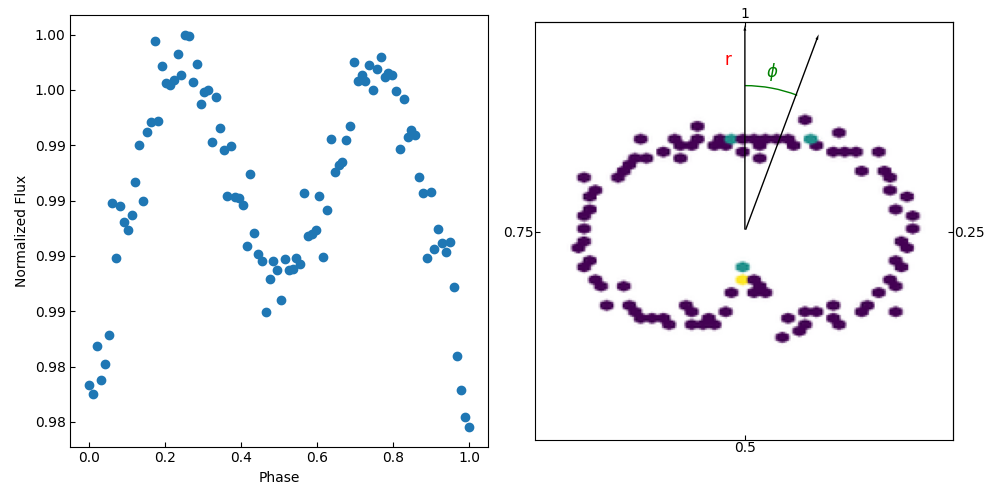}
    \includegraphics[width=0.4\textwidth]{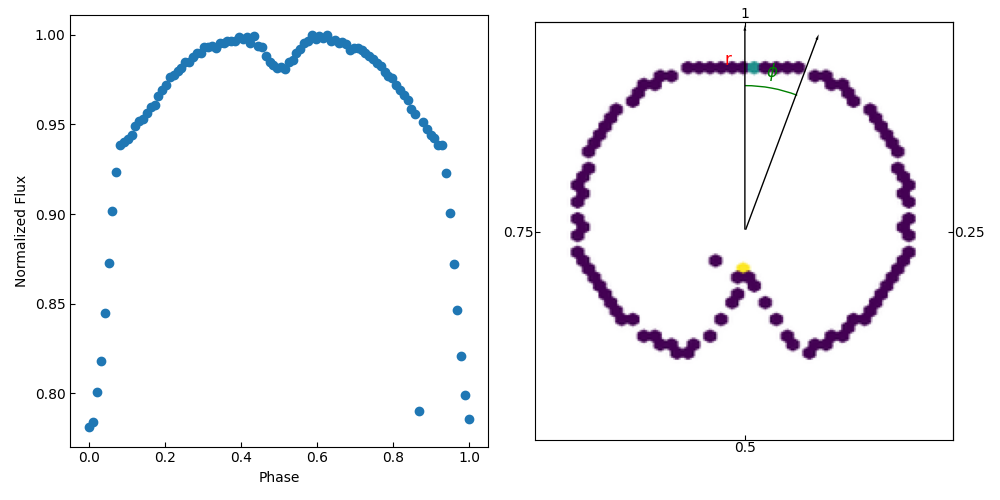}	
    
 \caption{Examples of image transformation for different classes of EBs. Top row: overcontact without spots, with spots. Low row: detached without spots and with spots.} 
    \label{example_transf}%
\end{figure}

 The breakthrough came from developing a transformation inspired by the Doppler tomography technique \citep{1988AdSpR...8b.127M}, which visualises emission line profiles observed over an orbital period to recover the line emissivity distribution in a two-dimensional space. In our approach, each light curve is transformed into polar coordinates. The flux values normalised on amplitude, ranging from 0.2 to 1, are mapped to the radius. In contrast, the phase values are converted to the polar angle (azimuth), measured clockwise from the vertical axis. In addition to the polar transformation, we also employed a hexbin \citep{84059e64-8132-3530-83f7-847a77216710} representation of the light curve points. This approach involves binning the data points onto a hexagonal grid and representing the number of points within each bin as a pixel value. The final visualization itself was also the result of a careful iterative process. We experimented with various colour palettes, `gridsize` parameters for the hexagonal grid, and different scaling options to achieve the best classification results presented in this paper. The final images were reshaped to 224×224 pixels, a standard input dimension for the CV models used. Examples of this transformation for different classes of EBs are shown on Figure~\ref{example_transf}.

To align the synthetic data more closely with observational characteristics and ensure our models would be robust, we subjected the  synthetic light curves to a series of augmentations. Photometric errors were simulated by adding a random value to each point, drawn from a normal distribution centred at zero. Crucially, the standard deviation ($\sigma$) of this distribution was specifically adjusted to emulate the typical photometric precision of the different surveys we aimed to replicate: $TESS$ and Gaia $G$ - 0.001, $I$ - 0.005, in normalised values. Furthermore, to account for other common real-world data issues, we introduced a small number of random outliers. 

To mimic the sparse nature of Gaia observations, we randomly removed data points, reducing the count to approximately 50, while for TESS and OGLE simulations, the 100-point resolution was maintained.

All these steps were essential for training models capable of generalising from idealized synthetic data to the noisy and often irregularly sampled data from actual astronomical surveys.

\section{Used datasets}
\label{sec:datasets}
\subsection{Training and validation datasets}
One of the challenges in classifying problems using CV is the need for sufficient input training data. The data typically comprises input features and corresponding labels. 

To create a training dataset, we need light curves that have been preprocessed (phase-folded, normalised, and binned to 100 points) and converted into an image representation. We have two possibilities for creating a training dataset to classify EBs. The first one is to use previously labelled data or analyse published light curves of EBs. However, we encounter several challenges, including insufficient numbers, possible system misclassification, poor light curve quality, and/or an unbalanced dataset. The second possibility is to create a synthetic set of light curves. As mentioned in Section~\ref{sec:eclipsingbinaries}, the light curve shape and the type of EB are determined by the physical properties of the components and the orbital parameters. This enables us to create a set of labelled input light curves for different combinations of parameters.

We used ELISa\footnote{https://github.com/mikecokina/elisa} code \citep{2021A&A...652A.156C}, which is a software package written in Python, to create this synthetic dataset. It was developed to assist users with star/binary modelling, light and radial-velocity curve analysis. The interface enables simple access to modelling capabilities (including spots and pulsations), rapid creation of synthetic light curves with desired parameters and extensive post-computed meta information, including the morphological classification of the system.

\begin{table}[]
    \centering
    \caption{Intervals of parameters used to create synthetic light curves for detached and overcontact binaries. See text for parameter definition.} 
    \vspace{0.3cm}
    \begin{tabular}{c|c c}
    \hline
         & Detached & Overcontact \\
         \hline
     $P$ [days]  &0.1-2.0  & 0.1-1.5 \\    
     $i$         & 30-90$^\circ$ & 30-90$^\circ$ \\
     $q$         & 0.01-2.0 & 0.01-2.0\\
     $\Omega_1$  & 2-6 & 2-5\\
     $\Omega_2$  & 2-6 & 2-5\\
     $T_1/T_2$  & 0.7-5 & 1-1.4\\
     \hline
     lon  & \multicolumn{2}{c}{30.0-330$^\circ$}\\
     lat  & \multicolumn{2}{c}{0-180$^\circ$}\\
     r    & \multicolumn{2}{c}{5.0-30$^\circ$}\\
     k    & \multicolumn{2}{c}{0.8} \\
     \hline
    \end{tabular}\\
    \label{tab:parameters}
\end{table}

The synthetic light curves of the EBs were created from values covering a wide range of stellar parameters. The used intervals of star parameters ($i$ - orbital inclination, $q$ - mass ratio, $\Omega_{1,2}$ - potentials of components, $T_1/T_2$ - effective temperature ratio) and spot parameters (lon - longitude, lat - latitude, r - diameter and k - temperature factor) for both types of systems are listed in Table~\ref{tab:parameters}. It should be noted that used intervals, mainly for detached binaries, do not cover the entire range of possibilities. The intervals are chosen to represent typical observed light curves. Random uniform distribution in these intervals was used to select parameters. The corresponding flux-normalised light curves were generated in the phase interval $<0.0-1.0>$ with 100 equally spaced phases for the selected set of parameters. A set of binary star parameters was discarded if the resulting surface gravity or effective temperature for any surface element fell beyond the valid interpolation range of the atmospheric models \citep{2003IAUS..210P.A20C}. When creating light curves with a spot, we randomly selected a star where we put the spot, chose spot parameters, and calculated the light curve in the same manner as above.

We chose to generate our synthetic light curves to emulate the distinct characteristics of three key surveys: TESS, OGLE, and Gaia. This selection allowed us to cover a wide range of observational scenarios. TESS represents high-cadence, high-precision photometry, providing a baseline of nearly ideal data. OGLE offers another high-cadence dataset but with typically larger photometric variance, enabling us to test the model's tolerance to higher levels of noise. Gaia introduces the critical challenge of sparse data, often with fewer than 50 irregularly sampled points.

We obtained four synthetic datasets for each passband (Gaia $G$, $I$, and $TESS$). We have 500\,000 simulated light curves for overcontact systems, while we have simulated 1\,000\,000 light curves for detached systems. These datasets can be used to create training and validation datasets with images used to train CV models (see Section~\ref{sec:models}). The public dataset description and the link to download it are provided in the GitHub repository\footnote{https://github.com/astroupjs/EBML}. 

\subsection{Testing datasets}
\label{testing}
The performance and quality of our models (see Section~\ref{sec:models}) were tested on actual observational data. We chose the OGLE (The Optical Gravitational Lensing Experiment) catalogue of EBs \citep{2016AcA....66..405S, 2016AcA....66..421P}. The great advantage of this data collection is that light curves are classified into "C" - contact and "NC" - non-contact systems, which corresponds to our classification scheme overcontact and detached systems, respectively. We randomly selected from the OGLE catalogue 200 systems and downloaded their corresponding light curves from the Gaia Eclipsing Binary Catalog (GEBC) \citep{2023A&A...674A..16M}. Moreover, we selected 90 overcontact stars from the largest catalog of individually studied
W UMa stars (WUMaCat) \citep{2021ApJS..254...10L} and 52 detached binaries from the Detached Eclipsing Binary Catalogue (DEBCat) \citep{2015ASPC..496..164S}. Those stars had their light curves in the Transiting Exoplanet Survey Satellite (TESS) archive \citep{2015JATIS...1a4003R} and GEBC. 

The light curves from OGLE were phase-folded using ephemerides available in the catalogue, binned to 100 points and normalised to maximum flux value. The light curves from TESS were modified using the same procedure, but ephemerides were taken from the WUMaCat and DEBCat catalogues. GEBC gives a period and uses the time of maximum flux as the reference epoch instead of the minimum, requiring a phase shift for proper folding. Light curves were normalised as in the previous cases, and no binning was applied because of the small number of data points, up to 50.

\section{Classification models and their performance}
\label{sec:models}

We developed a hierarchical classification system for binary stars, first distinguishing between overcontact and detached systems and then further classifying each type based on the presence or absence of starspots. This was achieved by fine-tuning three distinct classification models for each Gaia~$G$, $I$, and $TESS$ passband, resulting in a total of 18 models, 9 for each architecture. For clarity, we refer to these models as: \texttt{binary} (for the initial overcontact/detached classification), \texttt{detached\_spot}, and \texttt{overcontact\_spot}. Each of the nine models was trained on its own dataset of 100,000 images, with an equal number of images per class (50,000), and validated on a separate set of 10,000 images with equal class representation.

We used PyTorch\footnote{https://pytorch.org} for model implementation. We initialized a pre-trained \texttt{ResNet50} model from the \texttt{torchvision} module, adapting it for our binary classification by replacing the final layer with a two-neuron linear layer. A pre-trained \texttt{vit\_base\_patch16\_224} model from the \texttt{timm} library\footnote{https://pypi.org/project/timm} was used, setting \texttt{num\_classes=2}. 

All model trainings for this study were conducted on a performance local workstation. The system was equipped with an AMD Ryzen 9 series processor, 64 GB of RAM, and an NVIDIA GeForce RTX 4070 graphics processing unit (GPU) with 12 GB of VRAM. All deep learning models were implemented using the PyTorch framework, with GPU acceleration enabled by the CUDA toolkit to handle the computational demands of training. On this hardware, the training time per epoch was approximately 8 minutes for the \texttt{ResNet50} model and 40 minutes for the \texttt{vit\_base\_patch16\_224} model, reflecting the different computational complexities of the architectures.

\begin{figure}[t]
  \centering 
   \includegraphics[width=0.45\textwidth]{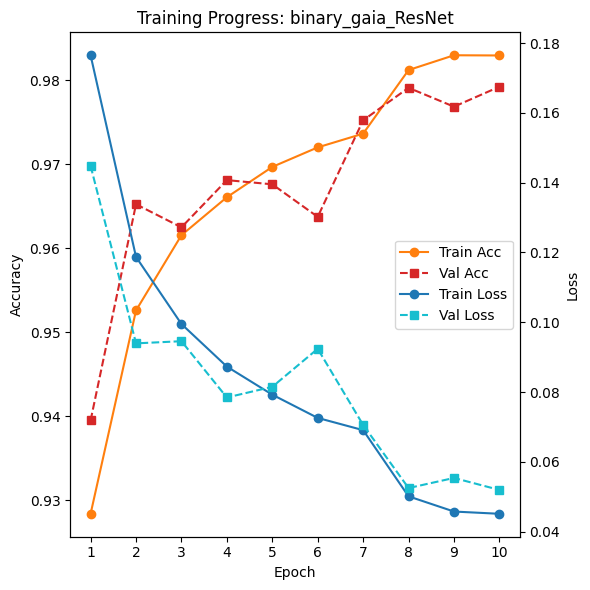}	
   \includegraphics[width=0.45\textwidth]{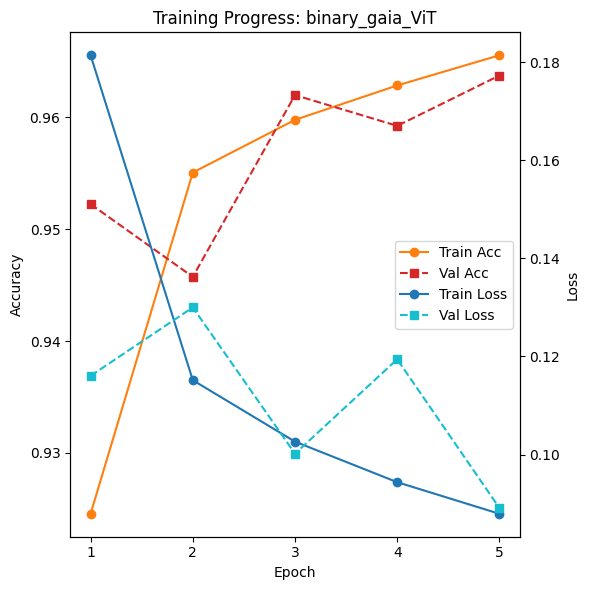}	
   
    \caption{Accuracy and loss training curves for Gaia $G$ passband $binary$ models.}
  \label{acc&loss}
\end{figure}

\begin{figure}[t]
    \centering 
    \includegraphics[width=0.15\textwidth]{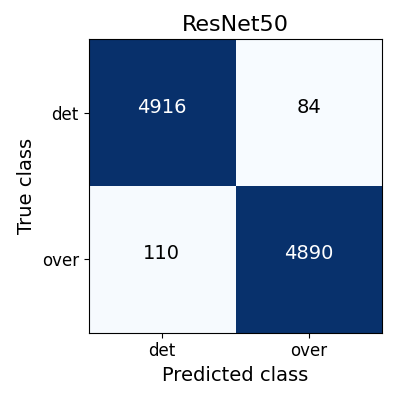}	
    \includegraphics[width=0.15\textwidth]{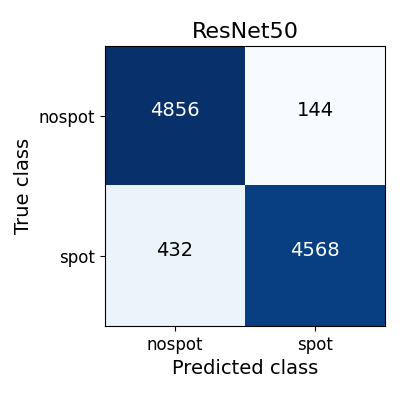}
    \includegraphics[width=0.15\textwidth]{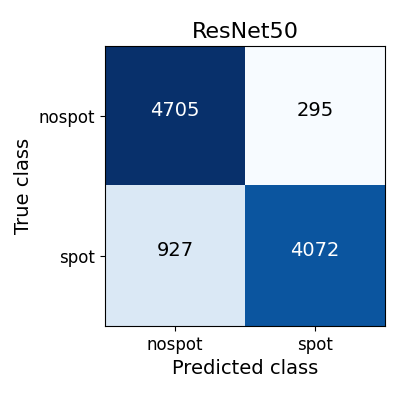}
    \includegraphics[width=0.15\textwidth]{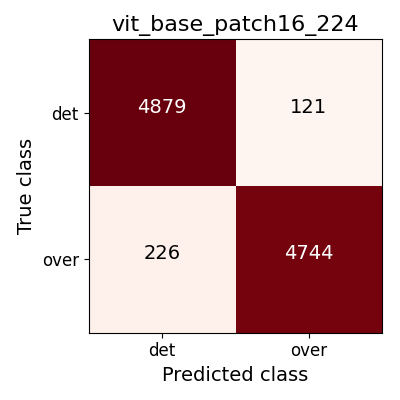}	
    \includegraphics[width=0.15\textwidth]{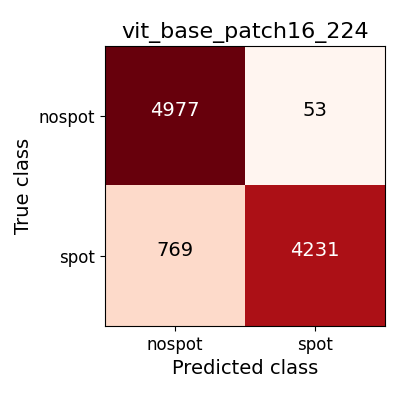}
    \includegraphics[width=0.15\textwidth]{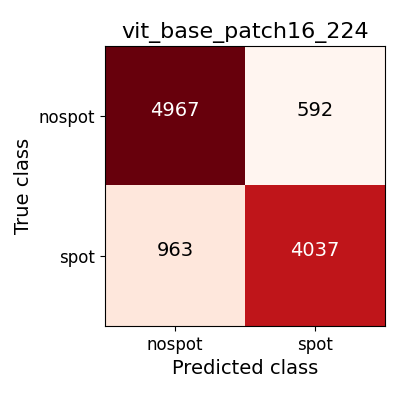}\\
    \includegraphics[width=0.15\textwidth]{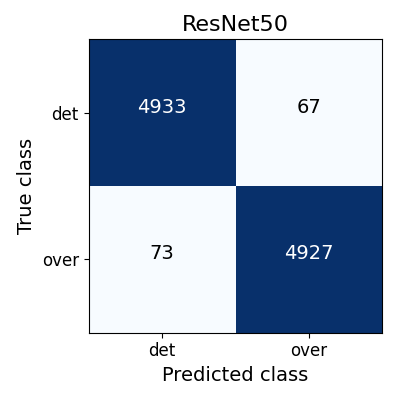}	
    \includegraphics[width=0.15\textwidth]{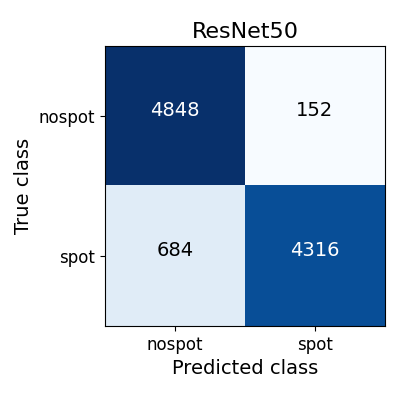}
    \includegraphics[width=0.15\textwidth]{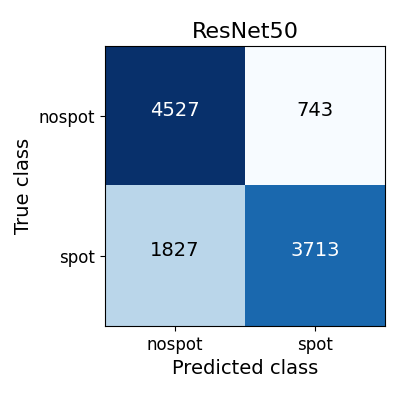}
    \includegraphics[width=0.15\textwidth]{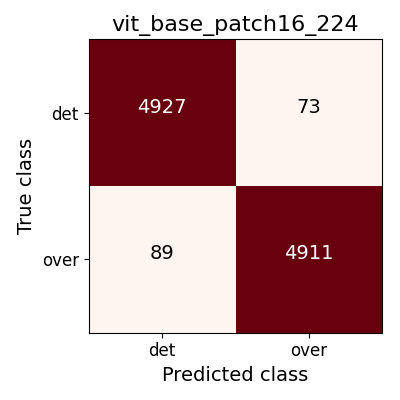}	
    \includegraphics[width=0.15\textwidth]{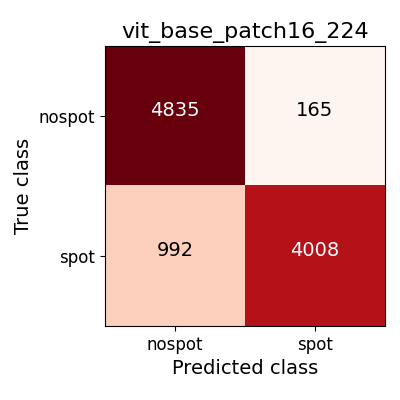}
    \includegraphics[width=0.15\textwidth]{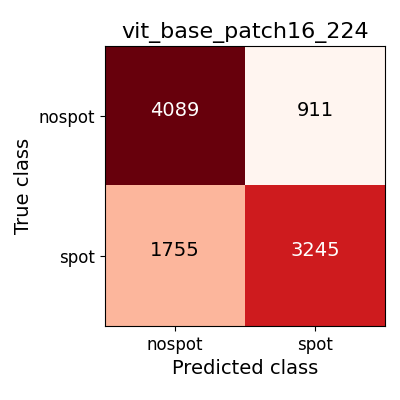}\\
    \includegraphics[width=0.15\textwidth]{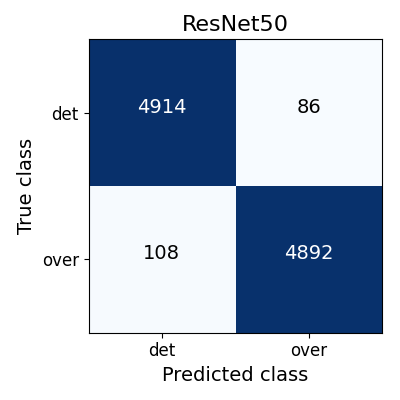}	
    \includegraphics[width=0.15\textwidth]{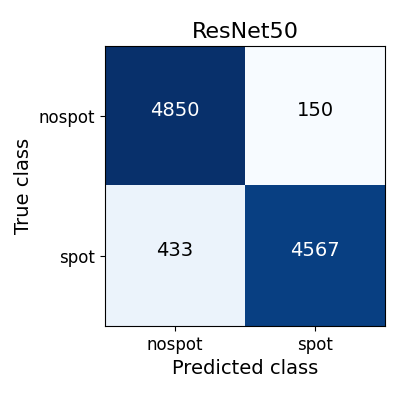}
    \includegraphics[width=0.15\textwidth]{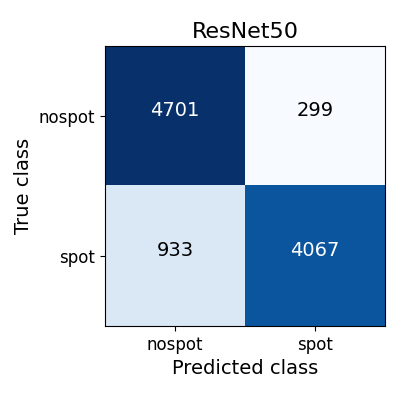}
    \includegraphics[width=0.15\textwidth]{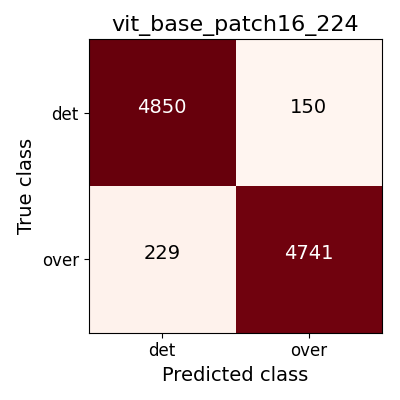}	
    \includegraphics[width=0.15\textwidth]{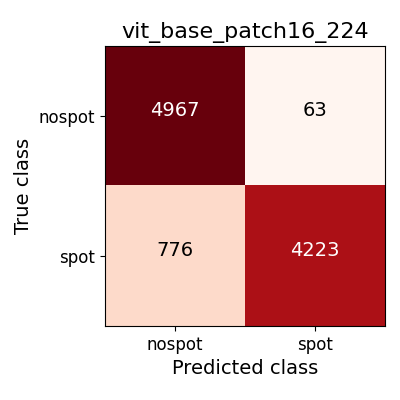}
    \includegraphics[width=0.15\textwidth]{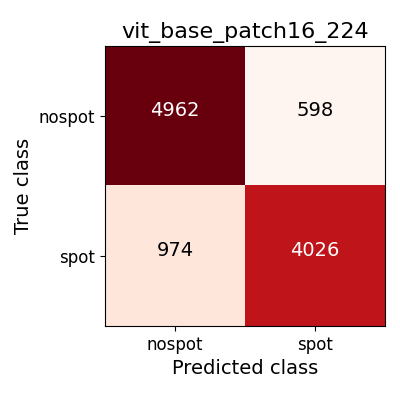}\\
    
 \caption{Confusion matrices of the models across different passbands, evaluated on the validation dataset. From left to right - \textit{binary, detached\_spot, overcontact\_spot}, from top to bottom - Gaia~$G$, $I$, and $TESS$.}  
    \label{conf_matrixes_resnet}%
\end{figure}

Training was carried out for 10 epochs for \texttt{ResNet50} and 5 epochs for \texttt{vit\_base\_patch16\_224}, using a batch size of 32 and a learning rate of 0.001, with loss and accuracy tracked at each epoch. Representative training curves are shown on Figure~\ref{acc&loss}; the full set of curves is available in the project's GitHub repository. The model accuracy on the validation datasets for the specific passbands is presented in Table~\ref{tab:accuracy} and confusion matrices on Figure~\ref{conf_matrixes_resnet}.

\subsection{Results on validation datasets}
\label{sec:valid_sample}

The validation accuracy of all \texttt{binary} classification models is consistent across the passbands, demonstrating strong performance of more than $96\%$. As we show further, this suggests the potential for a reliable application to real observational data. However, the performance of the \texttt{detached\_spot} models, while consistent across passbands, exhibits a decrease in accuracy compared to the \texttt{binary} models, less than $94\%$. The \texttt{overcontact\_spot} models demonstrate the lowest performance, particularly in the $I$ passband less than $77\%$. Analysis of the confusion matrices (Figure~\ref{conf_matrixes_resnet}) for the ResNet50 and ViT architectures reveals a systematic misclassification of spotted light curves as non-spotted. This trend is evident for overcontact binaries. This misclassification bias is probably attributable to the flux variations below the noise level in light curves influenced by small spots, which the models cannot detect. The reduced accuracy observed in the $I$ passband is probably a consequence of the lower signal-to-noise ratio in the training data for this band, which was designed to simulate the quality of the OGLE data. The increased noise level further diminishes the subtle photometric signatures of small spots, hindering model performance. 

\begin{table}[t]
    \centering
    \caption{Predicted classification accuracy of models for different passbands tested on the validation datasets.}
    \vspace{0.3cm}
    \begin{tabular}{lccc}
    \hline
     & Gaia $G$ & $I$ &  $TESS$  \\
    \hline
    \multicolumn{4}{l}{\bf ResNet50}\\
    \hline
     \texttt{binary}            &0.98& 0.98& 0.98 \\
     \texttt{detached\_spot}    &0.94& 0.92& 0.93 \\
     \texttt{overcontact\_spot} &0.88& 0.77& 0.87 \\
    \hline
    \multicolumn{4}{l}{\bf vit\_base\_patch16\_224}\\  
    \hline

        \cline{2-4}     
     \texttt{binary}            &0.96& 0.98& 0.98 \\
     \texttt{detached\_spot}    &0.92& 0.89& 0.91 \\
     \texttt{overcontact\_spot} &0.83& 0.75& 0.84 \\
     
    \hline
    \end{tabular}
   
    \label{tab:accuracy}
\end{table}

\subsection{Results on observational datasets}
\label{sec:test_sample}

\begin{figure*}[t]
    \centering 

        \includegraphics[width=0.32\textwidth]{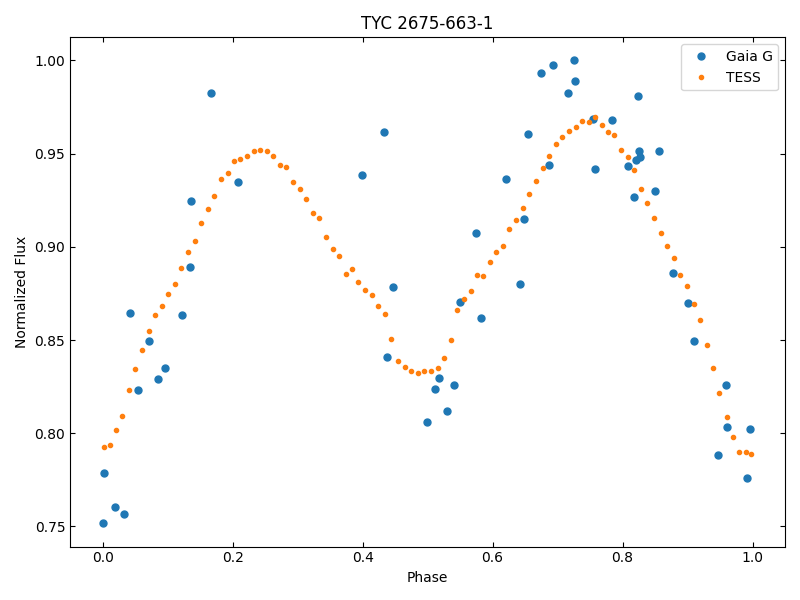} 
        \includegraphics[width=0.32\textwidth]{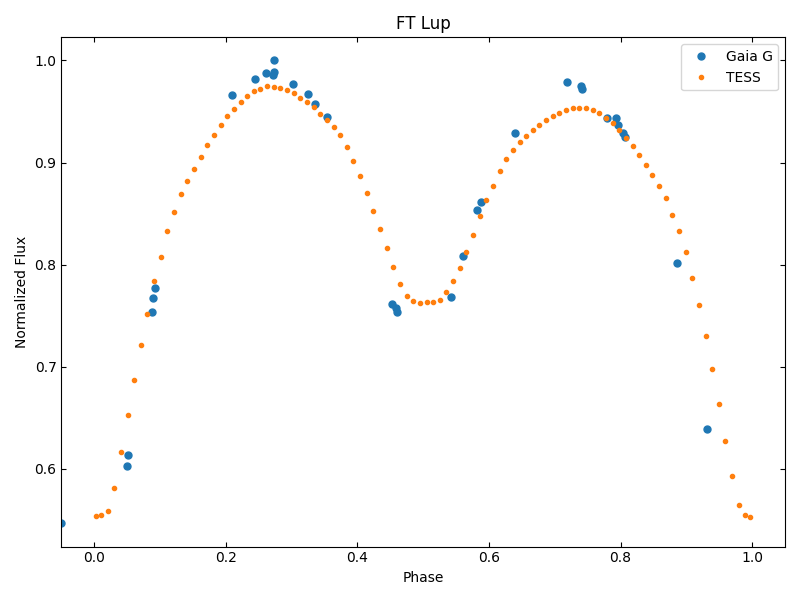}
        \includegraphics[width=0.32\textwidth]{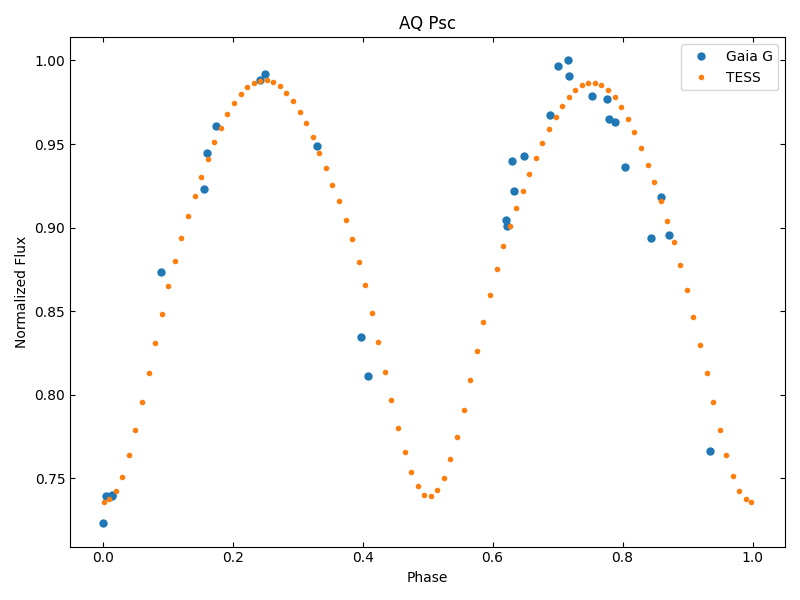} 
        \\
        \includegraphics[width=0.32\textwidth]{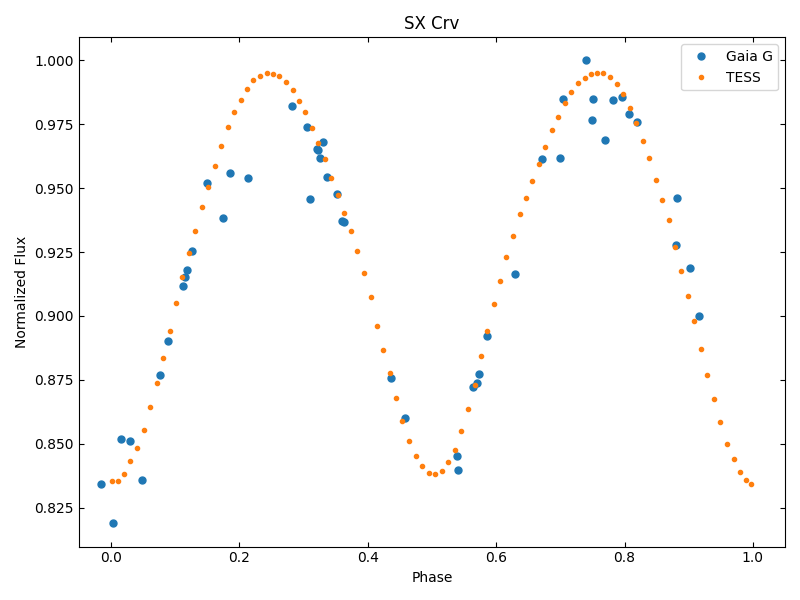} 
        \includegraphics[width=0.32\textwidth]{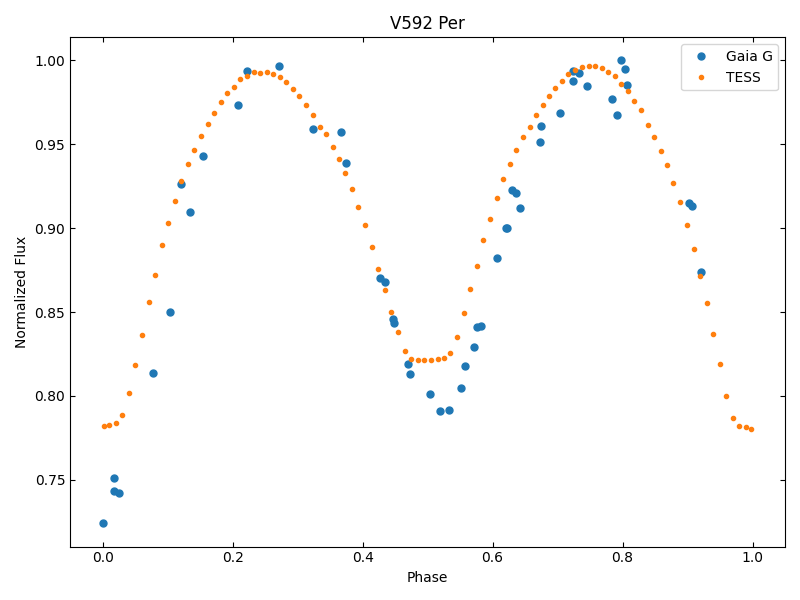} 
        \includegraphics[width=0.32\textwidth]{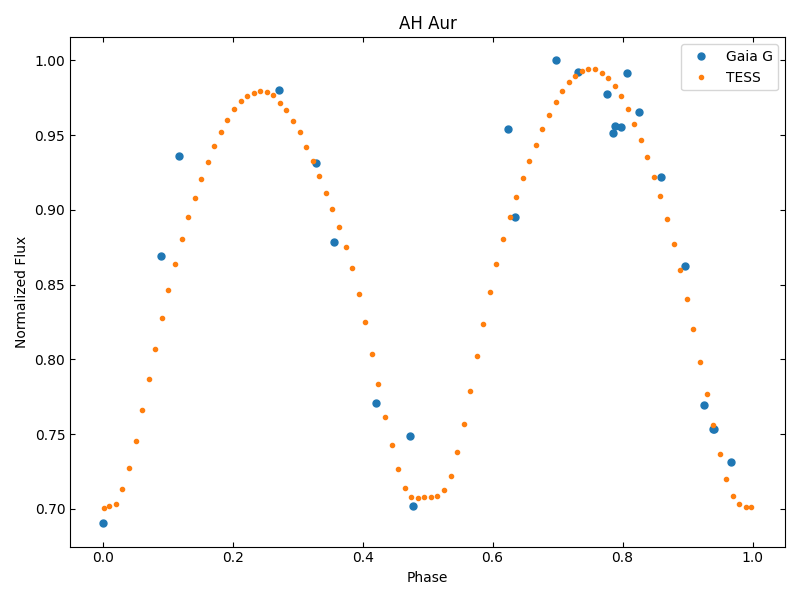} 
        \\
        \includegraphics[width=0.32\textwidth]{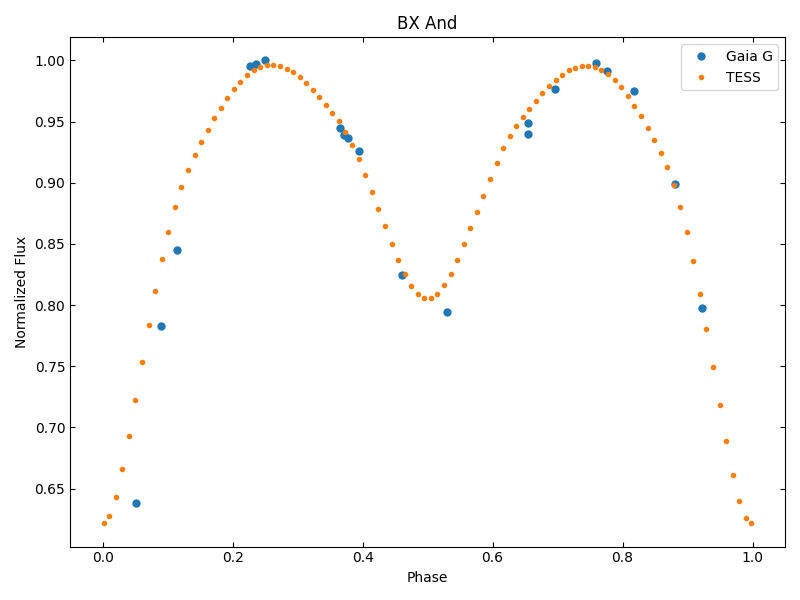} 
        \includegraphics[width=0.32\textwidth]{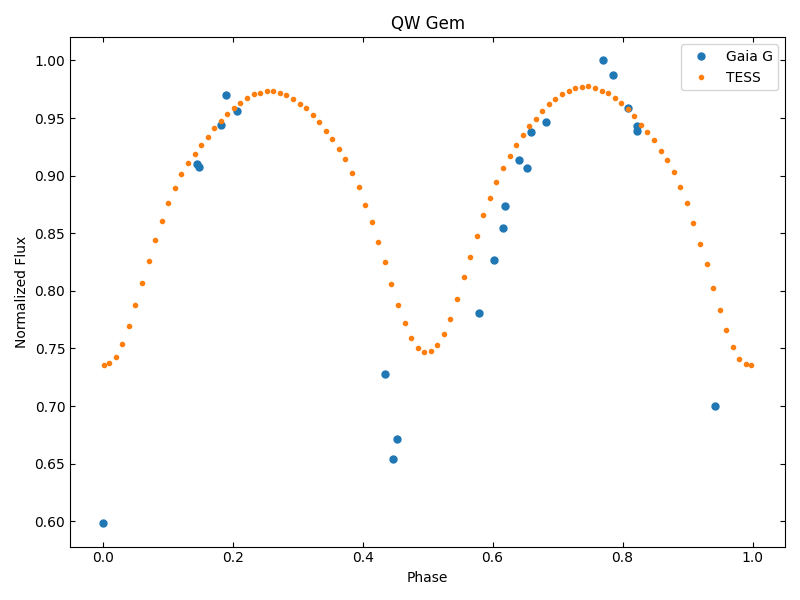} 
        \includegraphics[width=0.32\textwidth]{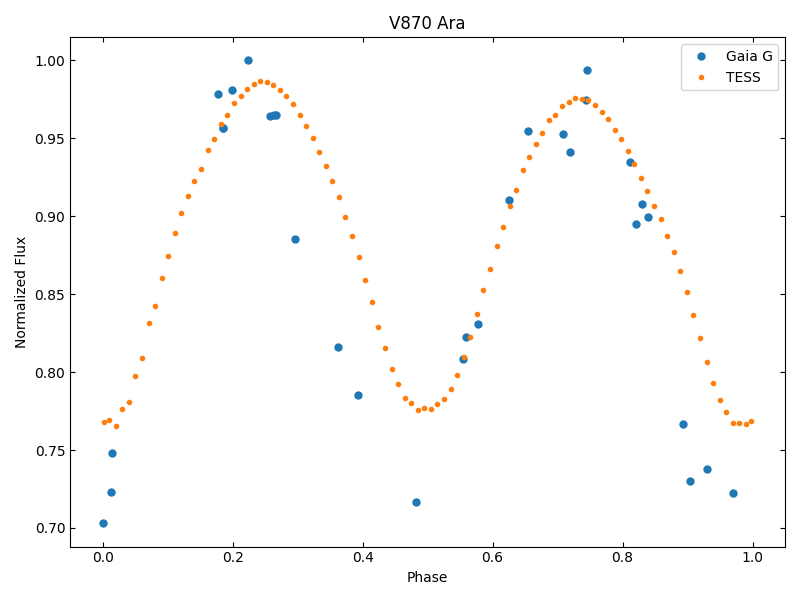} 

    \caption{Phased light curves (Gaia $G$ (blue) and $TESS$ (orange)) of the nine WUMaCat objects misclassified in the binary classification task. } 
    
    \label{fig:misclassified_lcs} 
\end{figure*}

We evaluated our models on the independent observational datasets described in Section \ref{testing}. Initially, the \texttt{binary} models were applied to TESS and Gaia light curves of known binary systems from the WUMaCat and DEBCat catalogues. Models applied to TESS light curves of both detached and overcontact systems achieved perfect classification accuracy. Similarly, detached systems observed in the Gaia $G$ band were classified without error. However, the ResNet50 and ViT models misclassified 3($3\%$) and 8($9\%$) overcontact systems in the Gaia $G$ band, respectively. Their TESS and Gaia light curves are shown on Figure~\ref{fig:misclassified_lcs}.

Two objects, \texttt{TYC 2675-663-1} and \texttt{FT Lup}, were misclassified by both ResNet50 and ViT. The light curve of the first object exhibits significant photometric noise, which may obscure the characteristic features required for correct classification by either model. The second system displays a notably shallow secondary eclipse. The less distinct shape, particularly the reduced depth of the secondary minimum, might deviate sufficiently from typical training examples, leading to misclassification. \texttt{AQ Psc} was misclassified by ResNet50 only. Its folded light curve contains substantial gaps in coverage, which can hinder the ResNet50 model's ability to recognise the complete light curve pattern from the generated image accurately. Misclassification of other objects by ViT is likely due to relatively low number of data points, resulting in a sparsely sampled light curve after phase folding. Sparse data might not provide enough detail for the ViT model's attention mechanisms. Light curves of two systems \texttt{SX Crv}, and \texttt{V592~Per} are visually well-defined and relatively clean. The reason for their misclassification by the ViT model is less obvious and might stem from more subtle differences between their features and the synthetic training data distribution, or specific sensitivities of the ViT architecture.

A separate randomly chosen test sample of 200 objects was compiled from the OGLE catalogue. This sample, having 124 detached and 76 overcontact systems (with corresponding $I$ and Gaia $G$ light curves for each object), utilised the existing OGLE classifications as ground truth. Performance on this OGLE sample yielded accuracies exceeding $94\%$ for both models in the Gaia $G$ band. In the $I$ band, models achieved accuracies greater than $95\%$ for overcontact systems but demonstrated significantly lower accuracy ($85\%$) for detached systems. Those and other metrics are presented in the Table \ref{tab:metrics_binary}

\begin{table}[t]
\centering
\caption{$Binary$ classification metrics for the test datasets. Upper section - the results for DEBCat detached objects as Negatives, WUMaCat overcontact objects as Positives. Lower section - 200 random OGLE-IV objects with the same encoding.} 
\vspace{0.3cm}
\begin{tabular}{c|c|c|c|c|c|c|c|c}

\label{tab:metrics_binary}

Model & Accuracy & Precision & Recall & F1 & TN & FP & FN & TP \\
\hline
\multicolumn{9}{c}{DEBcat and WUMaCat} \\
\hline
tess\_res & 1.00 & 1.0 & 1.00 & 1.00 & 52 & 0 & 0 & 90 \\
tess\_vit & 1.00 & 1.0 & 1.00 & 1.00 & 52 & 0 & 0 & 90 \\
gaia\_res & 0.98 & 1.0 & 0.97 & 0.98 & 52 & 0 & 3 & 87 \\
gaia\_vit & 0.94 & 1.0 & 0.91 & 0.95 & 52 & 0 & 8 & 82 \\
\hline
\multicolumn{9}{c}{OGLE-IV} \\
\hline
i\_res & 0.90 & 0.81 & 0.96 & 0.88 & 107 & 17 & 3 & 73 \\
i\_vit & 0.87 & 0.77 & 0.95 & 0.85 & 102 & 22 & 4 & 72 \\
gaia\_res & 0.97 & 0.95 & 0.97 & 0.96 & 120 & 4 & 2 & 74 \\
gaia\_vit & 0.94 & 0.90 & 0.96 & 0.93 & 116 & 8 & 3 & 73 \\
\end{tabular}
\end{table}

The performance of our models on the secondary task, detecting starspots, was evaluated using the combined DEBCat and WUMaCat samples. The classification metrics, presented in the Table~\ref{tab:spot_classification_results} , reveal a significant and systematic failure of our approach for this task.

For the overcontact systems, the results are exceptionally poor. With accuracy scores ranging from 0.49 to 0.60, were comparable to random guessing. This lack of diagnostic ability is confirmed by the area under the curve (AUC) values, which lay between 0.59 and 0.64, indicating almost no practical skill in distinguishing spotted from unspotted systems. The model for Gaia is a prime example of this failure; while its perfect recall (1.00) might seem impressive, it was achieved by classifying nearly every system as spotted, resulting in 42 false positives against only 2 true negatives. Such a model has no practical utility.

The performance on detached systems represents an even more comprehensive failure, particularly for the ViT architecture. Both Gaia and TESS models produced an F1-score of 0.00, meaning they failed to correctly identify a single spotted system (TP=0). The ResNet50 models performed no better in any meaningful sense, with extremely low F1-scores and recall values.

The AUC scores underscore this severe degradation in performance. The model for TESS achieved an AUC of 0.36, a result significantly worse than random chance, indicating that its predictions are actively misleading. The other models, with AUCs hovering around 0.51-0.54, confirm their inability to generalise from the training data for this task.

In summary, while the models were successful in the primary binary classification, they have proven to be entirely unreliable and unsuitable for the task of spot detection on real observational data. The subtle photometric signatures of spots are clearly being lost or misinterpreted, leading to a classification performance that is ineffective for practical application. 

 Comprehensive classification results for all models and test samples are available in our GitHub repository\footnote{https://github.com/astroupjs/EBML}. 

\begin{table}[t]

\centering
\caption{Spot detection classification metrics for objects from DEBcat and WUMaCat. Upper section - the results for overcontact objects with no spots as Negatives, with spots as Positives. Lower section - for detached objects with the same encoding.} 
\vspace{0.3cm}
\label{tab:spot_classification_results} 
\begin{tabular}{c|c|c|c|c|c|c|c|c|c}

Model & Acc & Prec & Recall & F1 & TN & FP & FN & TP & AUC\\
\hline
\multicolumn{9}{c}{Overcontact} \\
\hline
gaia\_resnet & 0.60 & 0.63 & 0.52 & 0.57 & 30 & 14 & 22 & 24 & 0.64\\
tess\_resnet & 0.50 & 0.51 & 0.83 & 0.63 & 7 & 37 & 8 & 38 & 0.59\\
gaia\_vit & 0.53 & 0.52 & 1.00 & 0.69 & 2 & 42 & 0 & 46 & 0.60\\
tess\_vit & 0.49 & 0.50 & 0.76 & 0.60 & 9 & 35 & 11 & 35 &0.6\\
\hline
\multicolumn{9}{c}{Detached} \\
\hline
gaia\_resnet & 0.63 & 0.42 & 0.29 & 0.34 & 28 & 7 & 12 & 5 & 0.53\\
tess\_resnet & 0.62 & 0.43 & 0.59 & 0.50 & 22 & 13 & 7 & 10 & 0.54\\
gaia\_vit & 0.67 & 0.00 & 0.00 & 0.00 & 35 & 0 & 17 & 0 & 0.51\\
tess\_vit & 0.63 & 0.00 & 0.00 & 0.00 & 33 & 2 & 17 & 0 & 0.36 \\
\hline
\end{tabular}
\end{table}

\section{Discussion}
\label{discussion}
Our study demonstrates that computer vision models (ResNet, ViT), applied to light curves transformed into specialised 2D images, effectively classify eclipsing binaries. A key strength of this work is the extensive testing performed on observational data for objects from the OGLE, DEBCat, and WUMaCat catalogues in different passbands, confirming the models' applicability beyond the synthetic data used for training.

The high accuracy achieved for binary classification (distinguishing detached vs. overcontact systems), exceeding 95-100\% on test sets (Section~\ref{sec:test_sample}), is consistent with the strong results obtained by \citet{2021A&C....3600488C} using sequence-based BiLSTM+CNN models on 1D vectors. This suggests our image-based CV approach offers a viable alternative methodology for this core classification task, potentially leveraging the robustness of CV models as discussed in Section~\ref{sec:comp_vis}. 

However, the secondary classification task, detecting starspots, proved significantly more challenging. Our models showed limitations in reliably identifying spots, especially for overcontact systems and in lower signal-to-noise data (e.g., $I$-band from $OGLE$), often misclassifying spotted systems as non-spotted or yielding high false positive rates on real data (Section~\ref{sec:test_sample}). 

This difficulty is notably consistent with the findings of \citet{2024CoSka..54b.167P}, who also attempted a four-class classification (including spots) using an LSTM+CNN approach and reported similar issues discerning spot signatures. This shared challenge highlights that robust, automated spot detection remains an open problem, likely requiring more diversity and realism within the synthetic training data, alongside advanced techniques to handle observational noise.

While our approach and the reference studies successfully utilized synthetic training data from ELISa \citep{2021A&A...652A.156C}, bridging the gap between synthetic and diverse real-world observations continues to be important. Our novel image transformation technique (Section~\ref{sec:transformation}) is a step in exploring alternative representations, but further work is needed to optimize feature extraction, especially for subtle effects like spots. Future efforts should focus on improving spot classification accuracy and rigorously comparing image-based versus sequence-based methods across identical, comprehensive datasets.

\section{Summary and conclusions}
In this work, we explored the application of CV techniques for the classification of eclipsing binary light curves. Our primary goal was to develop an automated method to distinguish between detached and overcontact systems, and further classify them based on the presence or absence of starspots. 

We employed two pre-trained deep learning models, ResNet50 (a CNN) and \textit{vit\_base\_patch16\_224} (a Vision Transformer), fine-tuning them for our specific classification tasks. A key aspect of our methodology involved generating large synthetic datasets of EB light curves using the ELISa code and transforming these phase-folded, normalised light curves into 224x224 pixel images. To mitigate over-fitting, we employed a multi-faceted strategy of extensive data augmentation, and our novel image representation which maps light curves into a polar coordinate and hexbin format. We created separate models for the primary binary classification (detached vs. overcontact) and the secondary spot detection task ($detached\_spot$, $overcontact\_spot$), training them on synthetic data designed to emulate real-world observational artefacts.

The performance evaluation yielded promising results, particularly for the initial binary classification task. On validation datasets derived from synthetic data, both ResNet50 and ViT models achieved high accuracy ($\>>96\%$) across Gaia $G$, $I$, and $TESS$ passbands. When tested on real observational data for objects from the DEBCat and WUMaCat catalogues ($TESS$ and Gaia $G$ bands), the binary classifiers performed exceptionally well. They achieved up to 100\% accuracy, with only minor misclassifications for some WUMaCat overcontact systems in Gaia $G$. Performance on a larger OGLE test sample also showed high accuracy ($\>>94\%$) for binary classification in Gaia $G$. 

In stark contrast, the models failed on the secondary task of automated spot detection. Performance on observational data was poor, often no better than random chance, with AUC scores staying between 0.36 and 0.64. The ViT models, in particular, were unable to detect any spotted detached systems, yielding F1-scores of 0.00. This demonstrates that our current models, despite their success in binary classification, are unsuitable for reliable spot detection in a practical setting.

In conclusion, our results demonstrate the significant potential of applying pre-trained computer vision models to classify eclipsing binary light curves, particularly for distinguishing detached and overcontact systems. Our novel light curve-to-image transformation technique appears effective for capturing relevant features. The high accuracy achieved in binary classification suggests this approach can be a valuable tool for analyzing the vast datasets from current and future surveys. However, our work also underscores that reliable, automated detection of subtle features like starspots remains an unsolved problem for these methods. Future work should focus on entirely new strategies to model and detect low-amplitude signals, as simple classification has proven inadequate.

\section*{Acknowledgements}
This research was supported by the Slovak Research and Development Agency under Contract No. APVV-20-0148. M.G was supported by EU NextGenerationEU through the Recovery and Resilience Plan for Slovakia under project No. 09I03-03-V04-00262. Y.M. was funded by the
EU NextGenerationEU through the Recovery and Resilience Plan for Slovakia under project No. 09I03-03-V01-00011 and internal 
grant of UPJ\v{S} No. VVGS 2024-309. P.G. was supported by the internal grant of UPJ\v{S} No. VVGS-2023-2784 and funded by the EU NextGenerationEU through the Recovery and Resilience Plan for Slovakia under project No. 09I03-03-V05-00008. M.V. acknowledges support of the project VEGA 2/0031/22.

\section*{Declaration of generative AI and AI-assisted technologies in the writing process.}
During the preparation of this work, the authors used Gemini 2.5 Pro to improve the readability and language of the manuscript. After using this tool, the authors reviewed and edited the content as needed and assume full responsibility for the content of the published article.

\bibliographystyle{elsarticle-harv} 
\bibliography{bibliography}

\end{document}